\documentclass[11pt,a4paper]{article}
\usepackage[hyperref]{emnlp2020}
\usepackage{times}
\usepackage{latexsym}

\usepackage{lipsum}
\usepackage{comment}

\usepackage[encapsulated]{CJK}

\usepackage{CJK}
\usepackage{hyperref}
\usepackage{amsmath}
\usepackage{graphicx}
\usepackage{multirow}
\usepackage{wrapfig}
\usepackage{subfig}
\usepackage[export]{adjustbox}

\usepackage{microtype}

\aclfinalcopy 

\title{Diving Deep into Context-Aware Neural Machine Translation}

\author{Jingjing Huo$^{1, 2}$ \quad Christian Herold$^{2}$ \quad Yingbo Gao$^{2}$ \quad Leonard Dahlmann$^{1}$\\ \textbf{Shahram Khadivi$^{1}$ \quad Hermann Ney$^{2}$} \\
$^{1}$eBay, Inc., Aachen, Germany\\
\texttt{\{jihuo, fdahlmann, skhadivi\}@ebay.com} \\
$^{2}$Human Language Technology and Pattern Recognition Group \\
RWTH Aachen University, Aachen, Germany \\
\texttt{\{surname\}@i6.informatik.rwth-aachen.de} 
}
\date{}

\begin{document}
\maketitle
\begin{abstract}

Context-aware neural machine translation (NMT) is a promising direction to improve the translation quality by making use of the additional context, e.g., document-level translation, or having meta-information. Although there exist various architectures and analyses, the effectiveness of different context-aware NMT models is not well explored yet.
This paper analyzes the performance of document-level NMT models on four diverse domains with a varied amount of parallel document-level bilingual data. We conduct a comprehensive set of experiments to investigate the impact of document-level NMT. 
We find that there is no single best approach to document-level NMT, but rather that different architectures come out on top on different tasks. Looking at task-specific problems, such as pronoun resolution or headline translation, we find improvements in the context-aware systems, even in cases where the corpus-level metrics like BLEU show no significant improvement. We also show that document-level back-translation significantly helps to compensate for the lack of document-level bi-texts. 

\includecomment{
Context-aware neural machine translation (NMT) is a promising direction for improving the translation quality having more context, e.g., document-level translation, or having meta-information. The goal is to enhance the translation of discourse phenomena and polysemous words.
This paper analyzes the performance of document-level NMT models with a varied amount of parallel document-level bilingual data. Including a diverse set of tasks, e.g., movie subtitles and e-commerce data, we conduct a comprehensive set of experiments to analyze and to learn the impact of document-level NMT. We show the document-level back-translation significantly helps to compensate for the lack of document-level bi-texts. 
}

\end{abstract}

\section{Introduction}
\label{sec:introduction}

Even though machine translation (MT) has greatly improved with the emergence of neural machine translation (NMT) \cite{NIPS2014_5346, DBLP:journals/corr/BahdanauCB14} and more recently the Transformer architecture \cite{vaswani2017attention}, there remain challenges which can not be solved by using sentence-level NMT systems.
Among other issues, this includes the problem of inter-sentential anaphora resolution \cite{guillou2018pronoun} or the consistent translation across a document \cite{laubli2018has}, for which the system inevitably needs document-level context information.

In recent years, many works have focused on changing existing NMT architectures to incorporate context information in the translation process \cite{tiedemann2017neural, bawden2017evaluating, voita2018context}.
However, often times results are reported only on very specific tasks (most commonly subtitle translation), making it difficult to assess the potential of the different methods in a more general setting.
This, together with the fact that big improvements are typically reported on low resource tasks, gives the impression that document-level NMT mostly improves due to regularization rather than from leveraging the additional context information.
In this work we want to give a more complete overview of the current state of document-level NMT by comparing various approaches on a variety of different tasks including an application-oriented E-commerce setting.
We discuss both, widely used performance metrics, as well as highly task-specific observations.

Another important aspect when talking about document-level NMT is the applicability in ``real life" settings.
There, when faced with a low resource data scenario, back-translation is an established way of greatly improving system performance \cite{sennrich2016improving}.
However, to the best of our knowledge, the effect of back-translation data obtained and used by context-aware models has never been explored before.
The main contributions of this paper are summarized below:
 \begin{itemize}
     \setlength\itemsep{-0.2em}
     \item We explore several existing context-aware architectures on four diverse machine translation tasks, consisting of different domains and data quantities.
     \item We examine the usage of context aware embeddings created by pre-trained monolingual models and study to what extent these embeddings can be simplified.
     \item We conduct corpus studies and extensive analysis on corpus specific phenomena like pronoun resolution or headline translation to give an interpretation of the potential improvements from leveraging context information.
     \item We study the effects of utilizing document-level monolingual data via back-translation and report significant improvements particularly for document-level NMT systems.
 \end{itemize}

\section{Related Works}
\label{sec:related_works}
The discourse- or document-level translation is a long-standing and unsolved topic in the machine translation community \cite{mitkov1999introduction, carpuat2009one, hardmeier2014discourse}.
Although neural machine translation \cite{NIPS2014_5346, DBLP:journals/corr/BahdanauCB14, vaswani2017attention} has recently become the dominant translation paradigm that provides superior performance, the independence between sentences is still the fundamental assumption taken for granted by most NMT systems. This means, that discourse-level phenomena between sentences such as pronominal reference, consistent lexical choice, and verbal tenses, etc. can not be addressed by these sentence-level NMT systems \cite{laubli2018has, guillou2018pronoun}.
The current NMT approaches tackling inter-sentential discourse phenomena can be roughly categorized into three aspects, augmenting NMT by
\begin{itemize}
    \setlength\itemsep{-0.2em}
    \item adding source-side context
    \item including both source- and target-side context
    \item utilizing source- and/or target-side document-level monolingual data
\end{itemize}

To include the source-side context, \citet{tiedemann2017neural} concatenate consecutive sentences as input to the NMT system, while \citet{jean2017does, bawden2017evaluating, zhang2018improving} use an additional encoder to extract contextual information from a few previous source-side sentences. 
These works only consider a local context, including a few previous sentences. Some researches seek to capture the global document context; \citet{wang2017exploiting} summarize the global context from all previous sentences in a document with a pre-trained hierarchical RNN and then use it for updating decoder states. Very recently, \citet{chen2020modeling} proposed a discourse structure-based encoder that takes account of the discourse structure information of the input document.

For adding additional target-side context, \citet{tiedemann2017neural, agrawal2018contextual} conduct multi-sentences decoding and observe only a minor improvement. 
\citet{maruf2017document} apply cache-based models to store vector representations for both source- and target-side context.
Similarly, \citet{tu2018learning} augment their NMT system with an external cache to memorize the translation history. 
\citet{miculicich2018document} integrate two hierarchical attention networks (HAN) \cite{yang2016hierarchical} in the NMT model to take account for source and target context. 
\citet{maruf2019selective} apply a hierarchical attention module on sentences and words in the context to select contextual information that is more relevant to the current sentence.

For incorporating document-level monolingual data from the source language, \citet{zhu2020incorporating} use BERT \cite{devlin2018bert} to model the source-side context and integrate it with the encoder and decoder of the NMT model. \citet{junczys2019microsoft} share the parameters of a BERT-style encoder trained on monolingual documents with the MT model.

To utilize the document-level monolingual data from the target language, \citet{junczys2019microsoft} also submit a system that trained on the combination of real and synthetic document-parallel data obtained by back-translation. However, they do not consider document-level back-translation. \citet{voita2019context} proposed a document-level post-editing system which is trained only using the monolingual document-level corpus.

Recently, there has been a tendency in the community to conclude that the context used in a context-aware MT model works as regularisation or noise generator.  \citet{kim2019and} compare several multi-encoders methods and claim that including this additional information can improve translation performance, but it is mostly due to the regularization effect rather than the contextual information. 
\citet{li2020does} also compare some context-aware architectures by replacing the real context with some random signal and show that random signals can achieve the same level improvement as the real context. However, it should be taken with a grain of salt since solving this task, along with the analysis, is quite challenging. There are many impact factors from the architecture, the data at hand, to the metric being used for evaluation.

One issue that can not be ignored in all discourse-related researches is the problem of evaluation. Since some discourse-level phenomena between sentences appear less frequently, although relevant, there is doubt if the metrics like BLEU score \cite{papineni2002bleu} can capture these complex relationships \cite{le2010aiding, hardmeier2010modelling}. To get more insights into the capacities dealing with discourse-level phenomena of their MT models, some researchers use more targeted evaluation scores \cite{wong2020contextual}, like the Accuracy of Pronoun Translation (APT) \citet{werlen2017validation}, or they evaluate their systems on some specific test suites that contain more and more complex discourse phenomena \cite{hardmeier2015pronoun, guillou2018pronoun, muller2018large, voita2019good}.

\section{Document-level NMT}
\label{sec:Document-level-NMT}

In this section, we first describe several commonly used context-aware NMT architectures and highlight the differences among them, largely following the work by \citet{kim2019and}.
Afterwards, we describe one radical attempt to represent the document-level context in one single embedding vector using BERT \cite{devlin2018bert}.
Finally, we explain our proposed paradigm to use document-level back-translation in detail.
Note that in this work, we consistently use Transformer \cite{vaswani2017attention} as our basic architecture and modify it accordingly.

\subsection{Context-Aware Architectures}
\label{sec:architectures}
\begin{figure}[ht]
    \centering
    \includegraphics[scale=0.6]{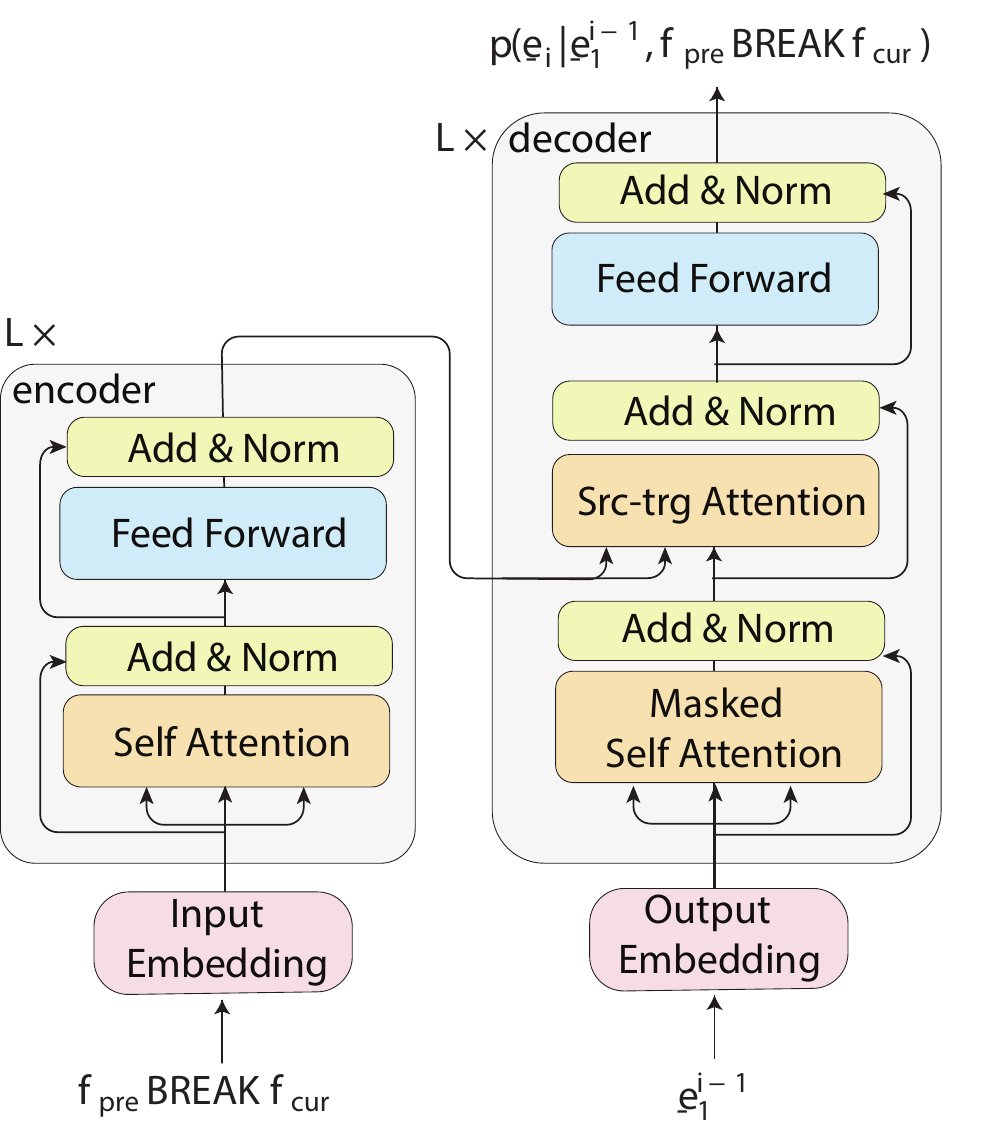}
    \caption{Single Encoder (2to2) approach only considering the one previous source sentence as context.}
    \label{fig:SE}
\end{figure}

Given a source sentence in a document to be translated, in order to exploit the source-side context from its previous sentences in the same document, a simple and straightforward technique is to concatenate these contextual sentences with the current source sentence \cite{tiedemann2017neural, agrawal2018contextual}. Similarly, if the previous and current target sentences are to be generated together, i.e. $\b{e}_1^{{I}} = e_1^{I_{\text{pre}}} \; \text{BREAK} \; e_1^{I{\text{cur}}}$, then the target-side context can also be considered by the model \cite{tiedemann2017neural}. Two additional special tokens are introduced to indicate the boundary between sentences and the beginning of a document, respectively. In this case, there is no modification of the model architecture itself, as seen in Figure \ref{fig:SE}.

\begin{figure}[ht]
    \centering
    \includegraphics[scale=0.5]{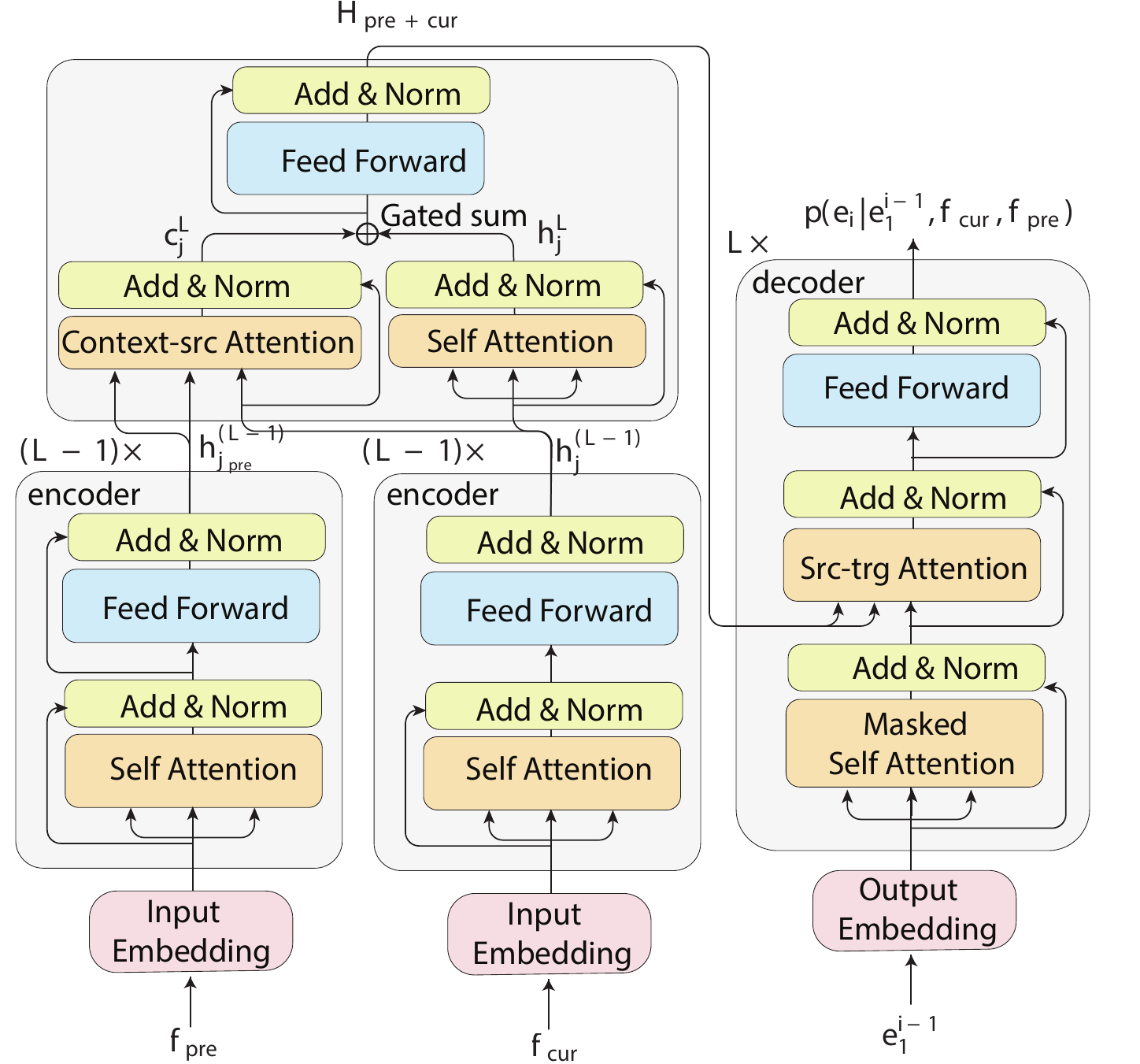}
    \caption{Multi-Encoders Out-side of decoder approach (Out.).}
    \label{fig:MEOD}
\end{figure}

An alternative way to model the source-side context is via an additional encoder, as shown in Figure \ref{fig:MEOD}. The previous sentence $f_{\text{pre}}$ is fed into an additional encoder to obtain the hidden representation of the source context sentence $h_{j_{\text{pre}}}^{L-1}$. At the last layer of the encoder, the source representation $h_j^{L-1}$ attends to $h_{j_{\text{pre}}}^{L-1}$ and outputs the combined hidden representation $c_j^{L}$ \cite{voita2018context}. Then, a gating mechanism \cite{bawden2017evaluating} between $h_j^{L}$ and $c_j^{L}$ is followed:
\begin{align}
  g_j & = \sigma (W_g [h_j^{L}, c_j^{L}] + b_g) \\
  o_j & = g_i \odot W_s h_j^{L} + (1 - g_i) \odot  W_c c_j^{L}
  \label{eq:AttentionGate}
\end{align}
We refer to this approach as ``Multi-Encoders (Out.)".

\begin{figure}[ht]
    \centering
    \includegraphics[scale=0.45]{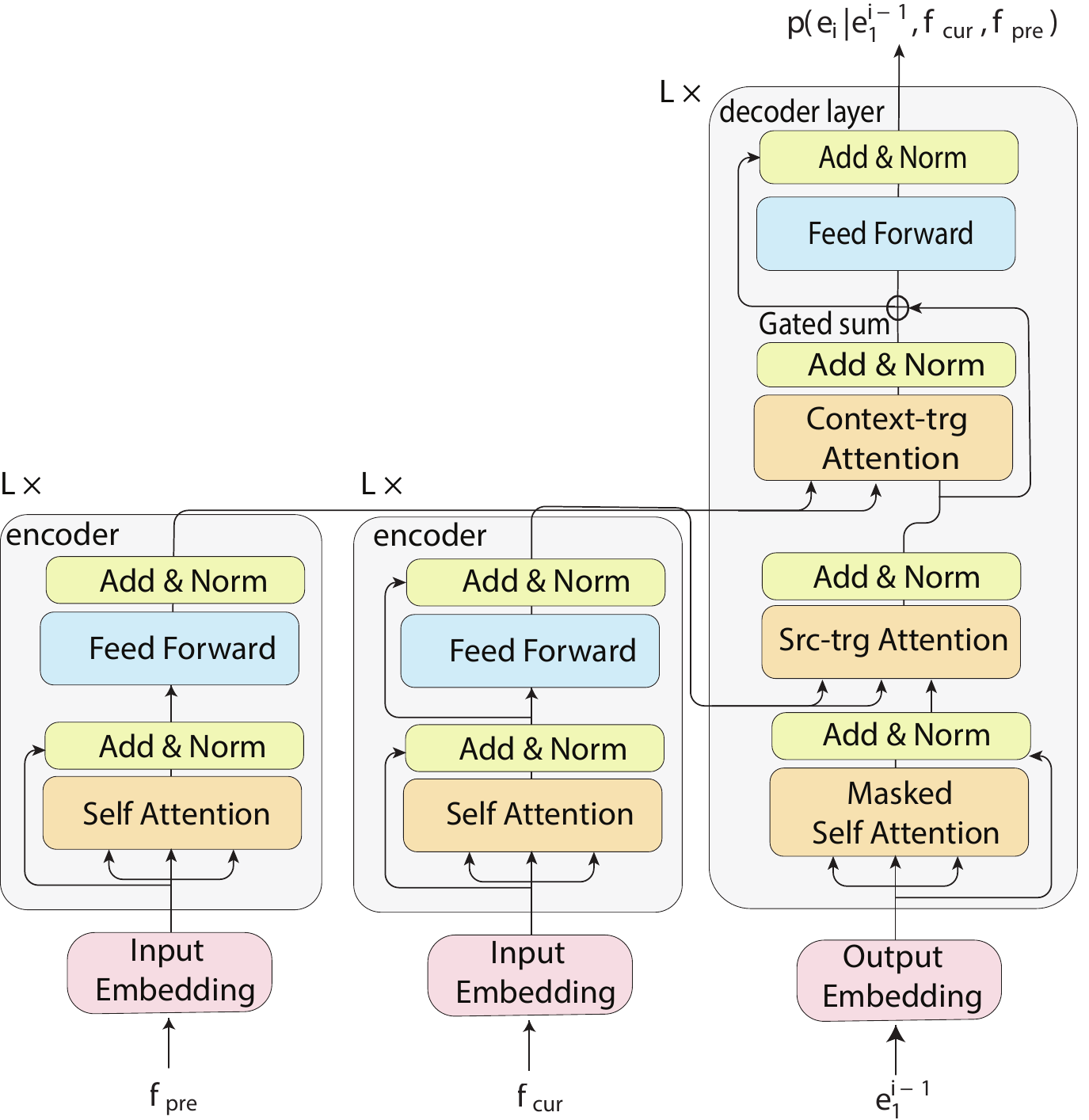}
    \caption{Multi-Encoders followed by attention components Inside of decoder Sequentially (In. Seq.).}
    \label{fig:MEIS}
\end{figure}

Another way to do the integration is to keep the representation of the current source sentence and the representation of the contexts separate and allow the decoder to have access to the context representations. Figure \ref{fig:MEIS} shows a sequential integration inside of the decoder, where the decoder firstly attends to the current source representation, then its output attends to the context representation \cite{zhang2018improving}. The same gating mechanism as in the Multi-Encoders (Out.) approach is used between the two attention outputs. We refer to this approach that uses multi-encoders followed by attention components inside of decoder sequentially as ``Multi-Encoders (In. Seq.)".

\begin{figure}[ht]
    \centering
    \includegraphics[scale=0.43]{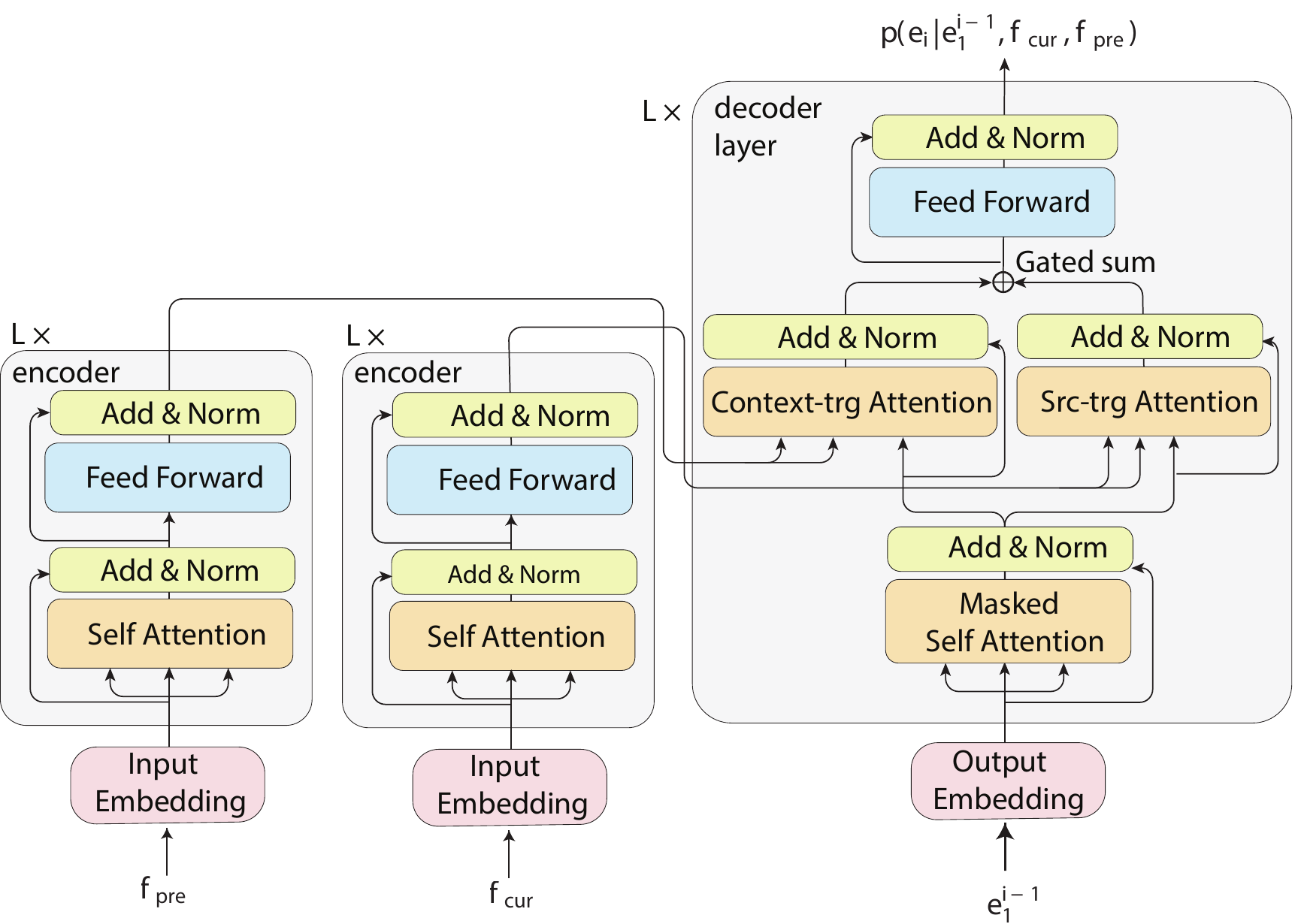}
    \caption{Multi-Encoders followed by attention components Inside of decoder in Parallel (In. Par.).}
    \label{fig:MEIP}
\end{figure}

Figure \ref{fig:MEIP} shows a parallel integration of the context inside of the decoder, where the decoder attends to the source and context representation in parallel and the outputs of them are combined again using a gating mechanism \cite{bawden2017evaluating}. In this paper, we call this approach using multiple encoders followed by attention components inside the decoder in parallel ``Multi-Encoders (In. Par.)''. 

In addition, we use ``WordEmb (In. Par.)" to refer to the approach that only uses word embeddings without any hidden layers to model the context and integrate it following the Multi-Encoders (In. Par.).

\begin{figure}[ht]
    \centering
    \includegraphics[scale=0.39]{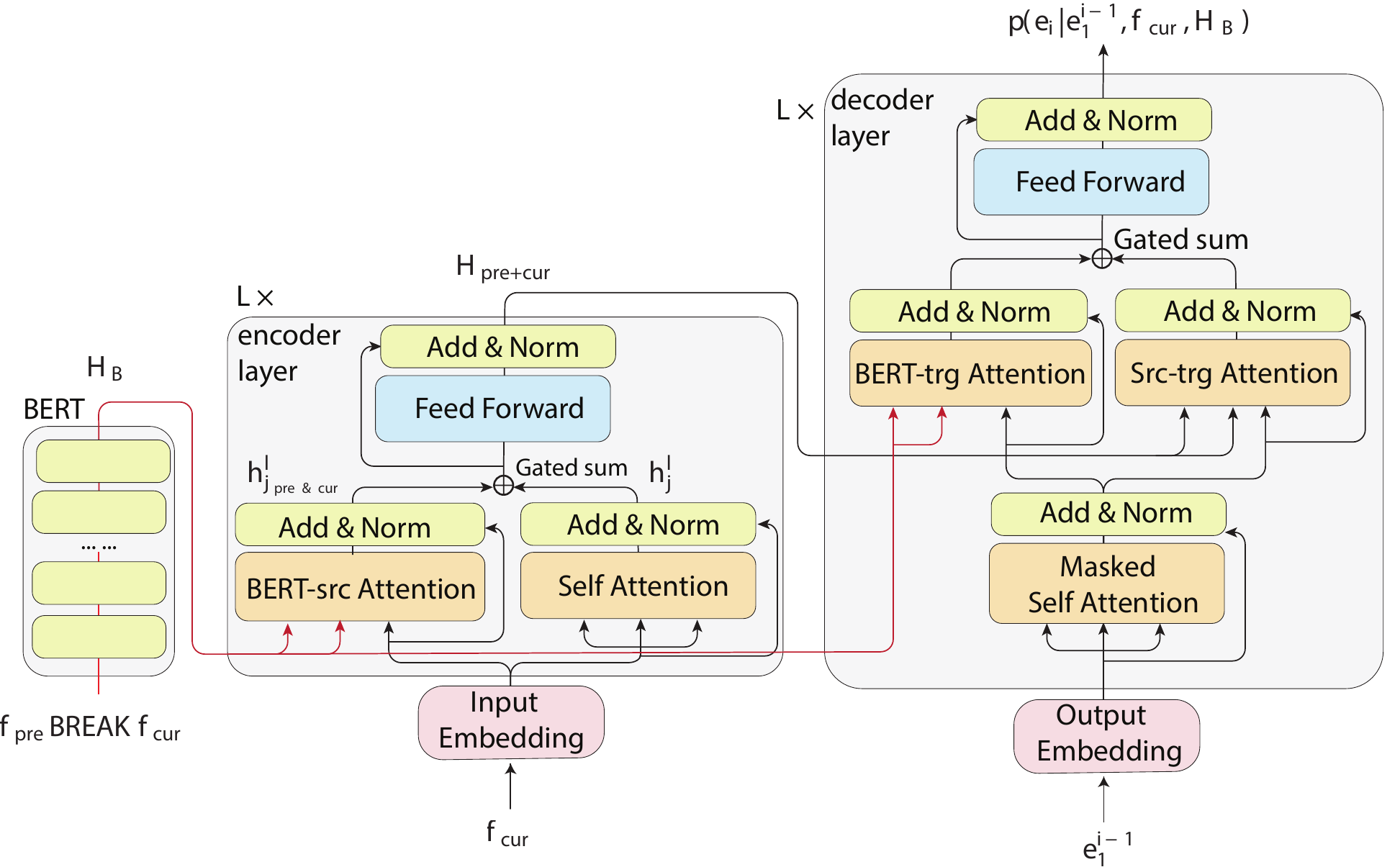}
    \caption{BERT sequence embeddings approach \cite{zhu2020incorporating}.}
    \label{fig:BERT-fused}
\end{figure}

Considering that a pre-trained model like ELMo \cite{peters2018deep} or BERT \cite{devlin2018bert} can capture rich representations of the input, it is apparent that one can also use it to model contextual information.
Figure \ref{fig:BERT-fused} shows the BERT-fused model proposed in \citet{zhu2020incorporating}, which uses a BERT encoder to obtain the BERT hidden representations $H_B$ on the concatenation of the context sentence $f_{\text{pre}}$ and the current source sentence $f_{\text{cur}}$. 
$H_B$ is further fused into each layer of the encoder and decoder of the NMT model using the attention mechanism to obtain the context representation. Instead of using the summation operation like in the original paper, we combine the context representation $h_{j_{\text{pre\&cur}}}^{l}$ and source representation $h_j^{l}$ with a gating mechanism on the encoder side. Similar operation for the integration on the decoder side is used. This approach corresponds to the ``BERT sequence embeddings (emb.)" approach in our main results in Table \ref{tab:main_results}.

\subsection{Single Embedding Vector as Context Representation}
\label{sec:singe_embedding}
The introduction of additional encoders or attention components in the approaches mentioned in Section \ref{sec:architectures} brings a large number of parameters, which is not always ideal. Further, we propose one radical attempt to summarize the document-level context into one single embedding vector. We average the embeddings in the context representation $H_B$ obtained by BERT to obtain one single mean-pooled embedding and then concatenate it with the word embeddings of the current source sentence along time axis (T-axis) or feature axis (F-axis). Besides, for the e-commerce dataset, we also apply a variant of BERT, which we call eBERT, that was trained with additional e-commerce item titles as supplement in-domain data.

\subsection{Document-level Back-translation}
\label{sec:doc-level-bt-description}
While there exist many works showing the improvements of context-aware systems, some major aspects are typically not covered - one of them being back-translation \cite{sennrich2016improving}.
Back-translation is an integral part when building the strongest possible systems and is currently the best way to include monolingual data in the training of a NMT system. It uses an inverse, target-to-source, MT model to generate synthetic source sentences given target-side monolingual sentences. There exists a series of works on this topic \cite{hoang2018iterative, burlot2019using, graca19:wmt}. However, the underlying inverse MT model used so far is mostly on the sentence level. 

In this work, we argue that back-translation could be even more crucial when training document-level NMT systems, since even for common language pairs like German-English we have very limited amounts of parallel document-level data while having an abundance of monolingual document-level data. 
In addition, except for using a sentence-level inverse NMT model, we also introduce a document-level inverse MT model to generate pseudo source documents given monolingual target-side documents. The intuition behind this approach is that we expect the document-level back-translation system to keep more inter-sentential discourse-phenomena in the synthetic source documents. If the back-translation system is merely on the sentence level, some discourse-phenomena, like consistent lexical choices, might not remain in the generated source documents. Losing this potentially large amount of discourse-phenomena is not beneficial for training a context-aware model. 

Since there is a large amount of document-level, monolingual in-domain data in the form of the NewsCrawl corpora, we conduct back-translation experiments on the WMT task. Here, we first train a baseline model and a context-aware model on the WMT news-commentary v14 in the reverse direction (De-En). We decide to use Multi-Encoders (In. Par.) as our inverse context-aware model, as it has the best performance on the WMT task. Then we sample 4.8M sentences pairs from news-docs2018 monolingual corpus,\footnote{\url{http://data.statmt.org/news-crawl/doc/de}} which contains 168K documents. 
Next, we use the inverse NMT models to translate them applying beam search with beam size 5 and concatenate the resulting bilingual synthetic data with the real documents in the news-commentary v14 dataset (En-De).   Finally, we compare the performance of a sentence-level baseline (En-De) and a context-aware model, Single Encoder (2to2), on both concatenated corpora. 
To our knowledge, this is the first attempt to explore the document-level back-translation data systematically (see Section \ref{sec:doc-level-back-translation}).

\section{Experiments}
\label{sec:experiments}
\subsection{Datasets}
\label{sec:datasets}

\begin{table*}[]
\centering
\begingroup
\setlength{\tabcolsep}{2pt} 
\renewcommand{\arraystretch}{1} 
\begin{tabular}{l|rrr|rrr|rrr|rrr}
\hline
\multicolumn{1}{c|}{} & \multicolumn{3}{c|}{IWSLT} & \multicolumn{3}{c|}{WMT} & \multicolumn{3}{c|}{OpenSubtitles} & \multicolumn{3}{c}{E-commerce data} \\ \hline
\# Sentences & 233K/ & 1.6K/ & 1.2K & 338K/ & 2.2K/ & 3.0K & 22.5M/ & 3.5K/ & 3.8K & 36K/ & 478/ & 1K \\
\# Running words & 4.7M/ & 31K/ & 22K & 8.3M/ & 47K/ & 68K & 188M/ & 30K/ & 30K & 596K/ & 12K/ & 26K \\
Avg. sentence length & 20/ & 20/ & 19 & 25/ & 22/ & 23 & 8/ & 9/ & 8 & 17/ & 25/ & 26 \\\hline
\end{tabular}
\endgroup
\caption{Training/development/test corpora statistics.}
\label{tab:data-statistics}
\end{table*}

We experiment with various parallel document-level datasets including IWSLT TED talk English-Italian,\footnote{\url{https://sites.google.com/site/iwsltevaluation2017}} WMT news-commentary v14 English-German,\footnote{\url{https://www.statmt.org/wmt18/translation-task.html}} OpenSubtitles \cite{lison2016opensubtitles2016} v2018 English-German\footnote{\url{http://opus.nlpl.eu/OpenSubtitles-v2018.php}} and an additional in-house e-commerce English-Chinese dataset. The test sets for the former two are the IWSLT 2017 test set and WMT newstest2018, respectively; for the latter two, we have created the dev and test sets ourselves by doing appropriate splits to the complete dataset.\footnote{We randomly sample complete documents from different years for dev and test set. The precise document IDs are: dev: \{1997/517700, 2002/696617, 2007/933906, 2012/2192989, 2017/6007584\}, test: \{1997/708495, 2002/257044, 2007/1036109, 2012/2322334, 2017/6190628\}} The data statistics of bilingual corpora used for fine-tuning context-aware models are summarized in Table \ref{tab:data-statistics}. In the IWSLT, WMT and OpenSubtitles datasets, there exists a boundary between documents. We first take them as sentence-level corpora to train the baseline and further fine-tune the context-aware system on them. 

The context-aware part of the e-commerce dataset is quite small and distinct from the other tasks: it does not contain documents or talks, but rather sentence-level item descriptions from an e-commerce website. As translation context, we provide the title of the item, instead of preceding sentences. Item descriptions and titles are user-provided, so they may contain ungrammatical sentences, spelling errors, and other noise. We give the title as context on the source-side, and we have reference translations only for the descriptions. In order to get a strong baseline, we additionally use a large sentence-level e-commerce dataset consisting of 6M sentence pairs (2.7M in-domain and 3.3M out-of-domain e-commerce) to train the baseline system, and then use it as initialization for fine-tuning on the context-aware e-commerce dataset. This dataset allows us to investigate context-aware NMT in a realistic scenario, in which the majority of training data does not have additional context.

To get a better insight into the model's performance for tackling the pronoun translation, we evaluate our models on two targeted test sets: one is ControPro for OpenSubtitles, the other is a coreference-focused test set for WMT. ControPro is introduced in \citet{muller2018large}, which is a contrastive test set extracted from OpenSubtitles with previous sentences as context. The source sentence has the English pronoun \textit{it} and three corresponding German translations containing German counterparts \textit{es}, \textit{sie}, \textit{er}, i.e., one of them is correct, and the other two are incorrect. The evaluation is done by counting the decisions that models rank the correct translation higher than the incorrect translations. In addition to using it in this way, we keep the source and the corresponding correct translation to form a standard test set containing 12K sentence pairs and measure the general translation quality on it.

\begin{table}[ht]
\centering
\begingroup
\setlength{\tabcolsep}{2pt} 
\renewcommand{\arraystretch}{1} 
\begin{tabular}{lrr}
\hline
 & \multicolumn{1}{c}{ControPro} & \multicolumn{1}{c}{Coreference} \\ \hline
\# Sentences &  12K  & 1.1K \\
\# Running words & 129K & 28K \\
Avg. sentence length & 11 & 26 \\ \hline
\end{tabular}
\endgroup
\caption{Two targeted-test sets: ControPro \cite{muller2018large} and coreference-focused test set extracted from WMT newstest 2008-2019 using NeuralCoref.}
\label{tab:data-statistics-targeted}
\end{table}
To create a targeted test set for WMT, we use an external tool called NeuralCoref\footnote{\url{https://github.com/huggingface/neuralcoref}}. We first apply this external tool to detect the coreference resolution in two consecutive sentences from newstest2008-2019, and then only keep the sentences where the coreference is resolved inter-sententially. This results in a targeted test set containing more inter-sentential discourse phenomena. The detailed statistics of these two targeted test sets are given in Table \ref{tab:data-statistics-targeted}.

All language pairs are preprocessed with the Moses tokenizer\footnote{\url{http://www.statmt.org/moses}} except for the Chinese corpus which is preprocessed with the chinese text segmentation tool ``jieba"\footnote{\url{https://github.com/fxsjy/jieba}}. We apply byte pair encoding \cite{sennrich2015neural} with 32k merge operations jointly for source and target languages. 

\subsection{Experimental setting}
All models are implemented in open-source toolkit OpenNMT \cite{klein2017opennmt}. For the sentence-level baseline system, we follow a 6-layer base Transformer model \cite{vaswani2017attention} and set the hidden size and embedding size as 512 and the dimension of the feed-forward layer as 2048. We use 8 heads for multi-head attention. For our context-aware models, we extend baseline system to include additional encoder with the same setting.
\begin{table*}[ht]
\centering
\begin{tabular}{l|r|r|r|r|r}
\hline
\multicolumn{1}{c|}{} & \multicolumn{1}{c|}{} & \multicolumn{1}{c|}{IWSLT} & \multicolumn{1}{c|}{WMT} & \multicolumn{1}{c|}{OpenSubtitles} & \multicolumn{1}{c}{E-commerce data} \\ \cline{3-6}
\multicolumn{1}{l|}{System} & \multicolumn{1}{c|}{Type} & \multicolumn{1}{c|}{BLEU{[}\%{]}} & \multicolumn{1}{c|}{BLEU{[}\%{]}} & \multicolumn{1}{c|}{BLEU{[}\%{]}} & \multicolumn{1}{c}{BLEU{[}\%{]}} \\ \hline
Baseline & N/A   & 31.6 & 28.4 & 37.3 & 33.7 \\ \hline
Single Encoder (2to1) & s & 31.7 & 28.3 & 37.5 & 32.8 \\
Single Encoder (3to1) & s &31.1 & 28.5 & 36.7 & N/A \\ \hline
Multi-Encoders (Out.) & s &31.3 & 28.6 & 37.6 & 34.0 \\
Multi-Encoders (In. Seq.) & s & 31.8 & 29.2 & 37.5 & \textbf{34.6} \\
Multi-Encoders (In. Par.) & s &32.2 & \textbf{30.1} & 37.5 & 34.2 \\
WordEmb (In. Par.)  & s &31.9 & 29.8 & 37.3 & 34.3 \\ \hline
Single Encoder (2to2) & s,t & 32.3 & 28.9 & \textbf{38.2} & N/A \\ \hline
BERT sequence emb. (e,d)& s,m & \textbf{32.8} & 29.0 & 37.4 & 34.0 \\
BERT sequence emb. (e)& s,m & 32.3 & 29.3 & 36.5 & 34.2 \\
BERT sequence emb. (d)& s,m  & 32.1 & 29.7 & 36.6 & 34.3 \\ \hline
BERT single emb. (T-axis)& s,m & 31.7 & 28.7 & 37.6 & 34.5 \\
eBERT single emb. (T-axis)& s,m  & N/A & N/A & N/A & 34.5 \\
BERT single emb. (F-axis)& s,m & 31.6 & 28.7 & 36.7 & 32.3 \\
\hline
\end{tabular}
\caption{Comparison of document-level architectures on different tasks. ``Type'' indicates whether the context used is from source(s)- or target(t)-side or if additional monolingual(m) data is included. ``e'' or ``d'' following the name of BERT sequence emb. approach indicates whether the context representation is fused on the encoder or decoder.}
\label{tab:main_results}
\end{table*}
In training, we use Adam optimizer \cite{kingma2014adam} or its variant Lazy Adam Optimizer for optimization and follow the learning rate schedule described in \cite{vaswani2017attention}. The learning rate scale factor and warm-up steps are different for different datasets.
In all our experiments, we share word embeddings over the source and the context. The context encoders are also initialized by the encoder of the sentence-level baseline. 

For automatic evaluation, we report case-sensitive sacreBLEU score \cite{post2018call} for all corpora except for e-commerce, on which the evaluation is done in Chinese character-level with case-insensitive sacreBLEU.

\subsection{Analysis}

\subsubsection{Performance in Terms of BLEU}
\label{sec:performanceinBLEU}
Table \ref{tab:main_results} shows the corpus-level BLEU-scores of all architectures on different tasks.
For the baseline as well as the ``source-side-only" systems we get similar results to \citet{kim2019and} on the IWSLT and WMT tasks, with Multi-Encoders (In. Par.) being the strongest architecture.
For the e-commerce data, Multi-Encoders (In. Seq.) performs slightly better.
Interestingly, with these architectures we do not see improvements on the much larger OpenSubtitles corpus.
This seems to confirm the suggestion of \citet{kim2019and} that these architectures work more as a regularization which is much more important for low resource tasks.

The Single Encoder (2to2) results in an improvement on all tasks excluding the e-commerce task, for which the method is not applicable due to the lack of target translation of the context (titles).
The improvements on the OpenSubtitles test set are comparable to what has been reported in the literature \cite{tiedemann2017neural} while the improvements on the other tasks are a bit smaller.
We notice that with this architecture, the improvements increase with decreasing average sentence length, which makes sense since it is known that the Transformer struggles with long input sequences \cite{rosendahl2019analysis}.
This seems also to be indicated by the deteriorating performance of the Single Encoder (3to1) system, which confirms the findings of \citet{agrawal2018contextual}.

Including context information through BERT sequence embeddings improves the performance on IWSLT, WMT and the e-commerce tasks but not on OpenSubtitles. The pre-trained BERT brings more (monolingual) data, which should again help primarily on the low resource tasks.
Contrary to the before mentioned approaches, the BERT single embedding approach does not significantly increase the number of free parameters, but it only works on the e-commerce task in our experiments.
This finding as well as the discrepancy between concatenating along the time or feature axis is discussed in detail in Section \ref{subsec:bert}.

While these findings are consistent with previous works, we find it to be insufficient to just rely on corpus-level BLEU scores to come to a conclusion about the usefulness of these approaches.
In the subsequent sections we discuss specific aspects of the translations which might be easily overlooked.
Furthermore we investigate the utilization of back-translation \cite{sennrich2016improving} for document-level systems, in an effort to compare these architectures in a more ``real-life" setting where back-translation is almost always used.


\subsubsection{Including BERT}
\label{subsec:bert}
\begin{table*}[ht]
\centering
\begin{tabular}{l|r|r|r|r|r|r}
\hline
\multirow{2}{*}{} & \multicolumn{2}{c|}{IWSLT} & \multicolumn{2}{c|}{WMT} & \multicolumn{2}{c}{E-commerce data} \\ \cline{2-7} 
System & \# tokens & \multicolumn{1}{l|}{BLEU[\%]} & \# tokens & \multicolumn{1}{l|}{BLEU[\%]} & \# tokens & \multicolumn{1}{l}{BLEU[\%]} \\ \hline
Reference & 19931 & - & 64276 & - & 40149 & - \\ \hline
Baseline & -226 & 31.6 & +1117 & 28.4 & -2672 & 33.7 \\ \hline
BERT single emb. (T-axis) & -66 & 31.7 & +879 & 28.7 & -2174 & 34.5 \\ \hline
Random emb. (T-axis) & +19 & 31.5 & +1557 & 28.7 & -2177 & 34.7 \\ \hline
\end{tabular}
\caption{Using different vectors for context representation. For the reference, the number of tokens stands for the total number of target tokens in the reference. In all consecutive lines, the number stands for the difference in the number of tokens compared to the reference.}
\label{tab:rnd_vec}
\end{table*}
When looking at the results in Table \ref{tab:main_results}, we see that using the embeddings produced by BERT yields some decent improvements on all tasks except for OpenSubtitles.
This might indicate that the improvements - at least in parts - come from the usage of additional data when training the BERT model rather than from an improved context representation.
A drawback when using the BERT system combination is the introduction of many additional parameters and calculations.
This can be drastically reduced when using a single vector extracted from BERT as described in Section \ref{sec:singe_embedding}.
However, the results of this approach are not significantly outperforming the baseline system on any tasks except for the e-commerce data.

Surprisingly, the eBERT does yield no further improvement over the BERT variant and the concatenation along the F-axis leads to a significant degradation in performance.
These two factors lead us to believe that the context information is not the decisive factor but something else.
To investigate this, we replaced the BERT-generated context vector with a random vector and compared the resulting BLEU scores which are shown in Table \ref{tab:rnd_vec}.

Depicted in this table are the BLEU score as well as the number of tokens in the respective hypothesis for the IWSLT, WMT and e-commerce tasks.
For replacing the real context vector we create the random vector by sampling from the uniform distribution.
Looking at the results, we see that our assumption is correct: the variant using a random vector yields the same improvements as the real context vector on the e-commerce task - even though it inhabits no relevant context information.

The reason behind this becomes clear when comparing the number of tokens produced in the hypotheses: On the e-commerce task we have a noticeable problem with under translation.
We argue that by increasing the length of the input sequence we inevitably increase the length of the output, leading to a longer hypothesis and consequently to a smaller brevity penalty when calculating BLEU.
This effect is not present for the other tasks at hand, since there we do not have a significant effect of under translation.
We note that similar results were obtained very recently by \citet{li2020does}, who also see improvements when replacing the context signal with random noise.
However, we conclude that the underlying effect is a different one, since we see no improvements when concatenating along the feature axis or when evaluating on a different task.
In conclusion, we argue that the improvements seen by using the BERT-embeddings for context information rather comes from additional data and other effects discussed in this section, rather than from the usage of actual context information.


\subsubsection{Better Headline Translation using Context}

In this section we discuss another unexpected effect of using context information in the translation, namely giving the system additional information about the nature of the input.
In the WMT task, both the train and test data consist of articles composed of a headline followed by a body of text, consisting of several sentences.
This means the only time the system has no context information at hand, is when translating the headline of an article.
We argue that the system can in fact use this information to distinguish whether the input sequence at hand is a headline or a real sentence and act accordingly.
Since a headline has a very distinguishable style compared with a complete sentence, this should lead to improvements in the translation quality.
To examine this hypothesis, we separate the WMT test set into two parts: One consisting only of headlines and the other one consisting only of body of texts.
We then evaluate the baseline system and our strongest document-level system (Multi-Encoders (In. Par.) for WMT) separately on both sets, The results can be seen in Table \ref{tab:title_vs_body}.


\begin{table}[ht]
\centering
\begin{tabular}{l|r}
\hline
System & \multicolumn{1}{c}{BLEU{[}\%{]}} \\ \hline
Baseline & 28.4\\
Doc-level & \textbf{30.1}\\ \hline
Baseline\_headlines & 19.9\\
Doc-level\_headlines & \textbf{24.4} \\ \hline
Baseline\_newsbody & 28.5 \\
Doc-level\_newsbody & \textbf{30.2} \\
\hline
\end{tabular}
\caption{System performance in terms of BLEU on headlines vs body of text for the WMT test set. The document-level system is Multi-Encoders (In. Par.).}
\label{tab:title_vs_body}
\end{table}

We see that the translations of both sets are improved when using the document-level setup.
However, the improvement on the headlines is much larger (+4.5\% BLEU) than on the body of text (+1.7\% BLEU).
When manually checking the hypotheses, we find that the baseline system often times tries to translate a headline as a ``complete" sentence (e.g. including a verb) while the document level system translates these in a much more consistent style.
This observation coincides with the fact that the baseline system shows severe signs of over-translation (on average 14.9\% more tokens than the reference) and the document-level system does not (-1.2\%).
We note that this effect is not responsible for the overall improvement in the corpus-level BLEU, since the ratio of headlines to text is very small (3.9\%).
This becomes clear when comparing the improvements on the body of text vs the complete test set - which is equal.
We conclude that this is another instance where the context improves the translation quality even if it is not immediately obvious. 


\subsubsection{Pronoun Resolution}
\label{sec:targeted-evaluation}
Testing the correct translation of pronouns is an established method to compare the context-awareness of document-level machine translation systems \cite{guillou2019findings, jean2017does, bawden2017evaluating, voita2018context, miculicich2018document, wong2020contextual}.
It can be argued that the ability of correctly translating inter-sentential pronouns not only depends on the architecture at hand but also on the data which the system is trained on. 
We decide to test the pronoun resolution capabilities of our systems in two different ways: First we are using an automatic metric for the accuracy of pronoun translation (APT) \cite{werlen2017validation} and second we use two targeted test sets described in Section \ref{sec:datasets}.
The results on OpenSubtitles and WMT can be found in Table \ref{tab:pronoun_res}.

\begin{table*}[ht]
\centering
\begingroup
\setlength{\tabcolsep}{5pt} 
\renewcommand{\arraystretch}{1} 
\begin{tabular}{l|rr|rrr||rr|rr}
\hline
 & \multicolumn{2}{c|}{OpenSubtitles} & \multicolumn{3}{c||}{ControPro}                              & \multicolumn{2}{c|}{WMT} & \multicolumn{2}{c}{Coreference test} \\ \cline{2-10} 
System & \multicolumn{1}{r}{BLEU} & \multicolumn{1}{r|}{APT} & \multicolumn{1}{r}{BLEU} & \multicolumn{1}{r}{APT} & \multicolumn{1}{r||}{corr. res.} & \multicolumn{1}{r}{BLEU} & \multicolumn{1}{r|}{APT} & \multicolumn{1}{r}{BLEU} & \multicolumn{1}{r}{APT} \\ \hline
Baseline & 37.3 & 52.8 & 30.5 & 35.4 & 48.7& 28.4 & 40.6  & 18.9  & 24.0 \\
Single Encoder (2to1) & 37.5 & 53.4 & \textbf{33.1} & 47.4 & 64.3 & 28.3 & 40.8 & 19.0  & 25.6 \\
Single Encoder (2to2) & \textbf{38.2} & \textbf{54.2} & \textbf{33.1} & \textbf{49.5} & \textbf{82.6} & \textbf{28.9} & \textbf{41.1} & \textbf{19.7} & \textbf{26.1}\\ \hline
\end{tabular}
\endgroup
\caption{Targeted evaluation for OpenSubtitles and WMT. All numbers are in percentage.}
\label{tab:pronoun_res}
\end{table*}

We calculate BLEU and APT scores on both the OpenSubtitles test set and ControPro test set (without contrastive translations) \cite{muller2018large}.
Furthermore we calculate the resolution accuracy on ControPro (with contrastive translations).
We compare the sentence-level baseline system with the best performing document-level system on this task - Single Encoder (2to2) as well as the Single Encoder (2to1) system.
Even though the latter does not significantly improve over the baseline on the OpenSubtitles test set, we find a significant increase in pronoun translation accuracy in terms of both evaluation methods.
The Single Encoder (2to2) system is even stronger in terms of pronoun translation, outperforming the baseline system by an impressive 33.9\% absolute accuracy on the targeted test set. When calculating BLEU on ControPro, the gap between the baseline and the document-level systems becomes significantly larger. The BLEU scores for the Single Encoder (2to2) and the Single Encoder (2to1) systems are equal.

When looking at the APT scores on WMT test set, the context-aware system does not provide much improvement. We assume the reason is that the portion of the potential improvement regarding inter-sentential pronoun resolution is quite small, having looked through this test set. The increased gap of APT score between the baseline system and the context-aware system on the coreference-focused test set confirms this assumption, as there are more inter-sentential coreference phenomena in this targeted test set. Note that the BLEU score gaps between the baseline and context-aware systems on both test sets are almost the same. 

All in all we can conclude that in this case the context information is helpful for a better translation, even though the effect might not be visible when just looking at corpus level BLEU.


\subsubsection{Document-level Back-translation}
\label{sec:doc-level-back-translation}
When dealing with document-level monolingual data, the question arises, whether a sentence-level back-translation system is sufficient to generate the synthetic data. In this section, we investigate the effect of the sentence-level back-translation data and document-level back-translation data on the baseline system as well as a context-aware system. The sentence-level baseline and context-aware model used to generate synthetic documents have 28.3\% BLEU and 29.7\% BLEU on the test set, respectively. The performance of the resulting En-De systems are summarized in Table \ref{tab:bt_exp}.


\begin{table}[ht]
\centering
\begin{tabular}{l|c|r}
\hline
BT-Data used & System & BLEU[\%] \\ \hline
- & Sent-level & 28.4 \\
 & Doc-level & 28.9  \\ \hline
Sent-level & Sent-level & 33.9 \\
 & Doc-level & 36.1 \\ \hline
Doc-level & Sent-level & 35.5\\
 & Doc-level & \textbf{36.5}  \\
 \hline
\end{tabular}
\caption{Including back-translation data to the WMT task. The architecture of the document-level system is the Single Encoder (2to2) approach.}
\label{tab:bt_exp}
\end{table}

When using the synthetic data generated by the sentence-level system we see a huge increase in performance for both systems (+5.5\% BLEU for the sentence-level system and +7.2\% BLEU for the document-level system).
A large increase in performance is to be expected since we increase the amount of data by roughly a factor of 8.
The systems trained on the synthetic data generated by the document-level system show even further improvements (+1.6\% BLEU for the sentence-level system and +0.4\% BLEU for the document-level system).
This might be in part due to the fact that the document-level back-translation system is stronger than the sentence-level one.

A very interesting observation is that the document-level system profits significantly more from the synthetic data in both scenarios.
This contradicts the proposition that document-level architectures function mainly as a form of regularization for low resource data-settings.
To the contrary we see an especially large gap in the case where we use only the sentence-level back-translation system for synthetic data generation.
We argue that the reason for this is, that the document-level system is more capable in recovering from errors made during the back-translation due to the context information.
For example a wrongly translated pronoun on the source side will definitely lead the sentence-level system astray, but the document-level one might still recover when the context is correct.
This assumptions is also supported by the fact that the gap between sentence-level and document-level system gets smaller when using synthetic data generated by the document-level system, since we assume less such errors get made by this system.



\section{Conclusion}
\label{sec:conclusion}

In this work, we give a comprehensive comparison of current approaches to document-level NMT.
To draw meaningful conclusions, we report results for standard NMT metrics on four diverse tasks - differing in the domain and the data size.
We find that there is no single best approach to document-level NMT, but rather that different architectures work the best on various tasks.
Looking at task-specific problems, such as pronoun resolution or headline translation, we find improvements in the context-aware systems, which is not visible in the corpus-level metric scores.

We also investigate methods to include document-level monolingual data on both source (using pre-trained embeddings) and target (using back-translation) sides.
We argue that the performance improvements from the pre-trained encoder predominantly come from increased training data and other task-specific phenomena unrelated to actual context information utilization.
Regarding back-translation, we find that document-level systems seem to benefit more from synthetically generated data than their sentence-level counterparts.
We discuss that this is because document-level systems are more robust to sentence-level noise.

We plan to expand our experiments to incorporate document-level monolingual data on both source and target sides.
This makes sense just by looking at the data conditions of almost every task: document-level parallel data is scarce, but document-level monolingual data is abundant.

\includecomment{
In this work we give a comprehensive comparison of current approaches to document-level NMT.
In order to draw meaningful conclusions, we report results for standard NMT metrics on four diverse tasks - differing both in domain as well as in data size.
We find that there is not a single best approach to document-level NMT, but rather that different architectures work best on different tasks.
Looking at task-specific problems, such as pronoun resolution or headline translation, we find improvements in the context-aware systems which can not be seen in the corpus-level metric scores.
We also investigate methods to include document-level monolingual data on both source (using pre-trained embeddings) and target (using back-translation) side.
We argue that the performance improvements from the pre-trained encoder predominantly come from increased training data and other task-specific phenomena unrelated to actual context information utilization.
Regarding back-translation, we find that document-level systems seem to benefit more from synthetically generated data than their sentence level counterparts.
We argue that this is because document-level systems are more robust to sentence-level noise.
We plan to expand our experiments regarding the incorporation of document-level monolingual data on both source and target side.
This makes sense just by looking at the data conditions of almost every task: document-level parallel data is very rare but document-level monolingual data is easy to come by.
}

\section*{Acknowledgments}


\begin{center}
    \includegraphics[width=0.14\textwidth, valign=m]{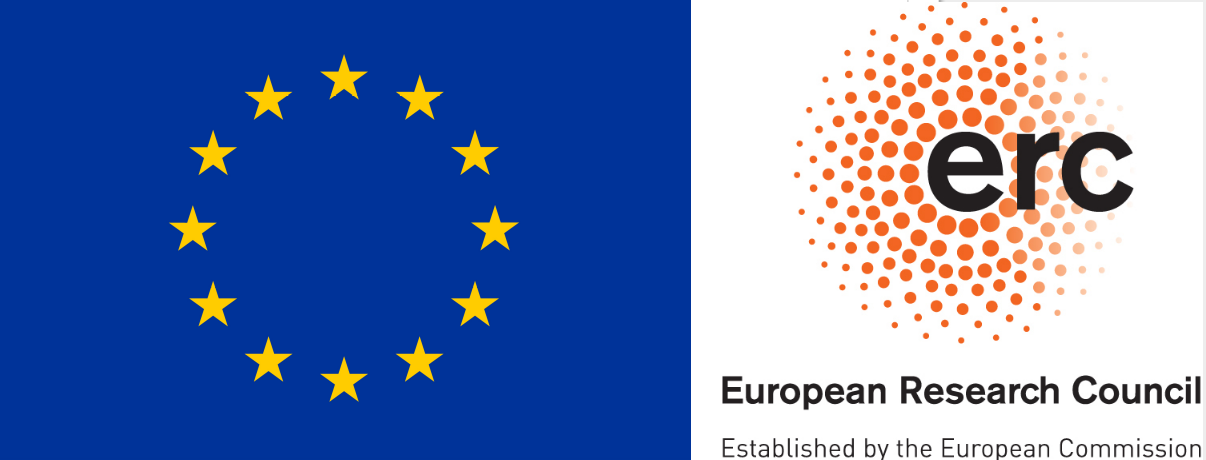}
    \hspace{2mm}
    \includegraphics[width=0.14\textwidth, valign=m]{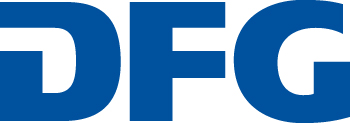}
    \hspace{2mm}
    \includegraphics[width=0.14\textwidth, valign=m]{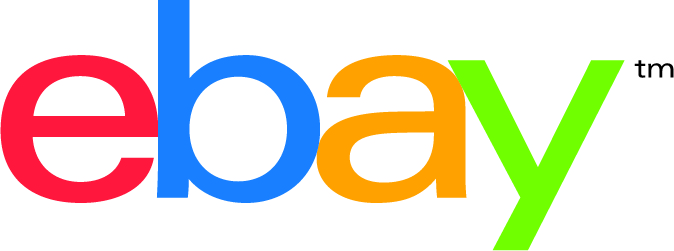}
\end{center}


Christian Herold and Yingbo Gao have received funding from the European Research Council (ERC) (under the European Union's Horizon 2020 research and innovation programme, grant agreement No 694537, project ``SEQCLAS") and the Deutsche Forschungsgemeinschaft (DFG; grant agreement NE 572/8-1, project ``CoreTec") and eBay Inc.
The work reflects only the authors' views and none of the funding agencies
is responsible for any use that may be made of the information it contains.

\bibliography{anthology,emnlp2020}
\bibliographystyle{acl_natbib}

\clearpage
\appendix
\section{Appendices}
\label{sec:appendix}
Here we provide some translation examples of sentence-level baseline and context-aware model for OpenSubtitles and WMT news-headlines. 

\begin{table}[ht]   
    \centering
    \small 
    \begin{tabular}{@{} l | l @{}}
    \hline
    Context & Make yourself comfortable. \\
    Source & I will check in on you later. \\
    Reference & Ich sehe später nach Ihnen.  \\
    \hline
    Sent-level pre & Machen \textbf{Sie} es sich bequem. \\
    Sent-level cur & Ich sehe später nach \textbf{dir}. \\
    Doc-level pre & Machen \textbf{Sie} es sich bequem. \\
    Doc-level cur & Ich sehe später nach \textbf{Ihnen}.\\
    \hline
    \hline 
    Context & He is never spent the night with a woman\\
    & before. \\
    Source & A woman? \\
    Reference & Einer Frau?  \\
    \hline
    Sent-level pre & Er hat noch nie die Nacht mit \textbf{einer} Frau \\
    & verbracht. \\
    Sent-level cur & \textbf{Eine} Frau? \\
    Doc-level pre & Er hat noch nie mit \textbf{einer} Frau geschlafen. \\
    Doc-level cur & \textbf{Einer} Frau?\\
    \hline
    \hline
    Context & I curse the fucking day you were born! \\
    Source & I curse it! \\
    Reference & Ich verfluche ihn!  \\
    \hline
    Sent-level pre &  Ich verfluche \textbf{den} Tag, an dem du  geboren \\ 
    & wurdest! \\
    Sent-level cur &  Ich verfluche \textbf{es}! \\
    Doc-level pre & Ich verfluche \textbf{den} Tag, an dem du geboren \\ &  wurdest! \\
    Doc-level cur & Ich verfluche \textbf{ihn}!\\
    \hline
    \hline
    Context & Where is what? I do not even know what \\
    & you are looking for.\\
    Source & Yes, you do.\\ 
    Reference & - Doch, das wissen Sie.  \\
    \hline
    Sent-level pre &  Ich \textbf{weiß} nicht mal, wonach du suchst.\\
    Sent-level cur &  Doch, das \textbf{tuest} du. \\
    Doc-level pre & Ich \textbf{weiß} nicht mal, wonach Sie suchen. \\
    Doc-level cur & Doch, das \textbf{wissen} Sie.\\
    \hline
    \end{tabular}
    \caption{Positive example translations in OpenSubtitles.}
    \label{tab:example}
\end{table}
Table \ref{tab:example} provides some positive cases that the best performing context-aware model, Single Encoder (2to2), utilizes the source- and target- side context to make the translations more consistent, while Table \ref{tab:example-opensubtitles-na} shows one improved example by the context-aware model where the context is not clearly useful: 
 \begin{table}[ht]  
        \centering
        \small 
        \begin{tabular}{@{} l | l @{}}
        \hline 
        Context &   Someone must have really wanted her ring.\\ 
        Source & Her nail beds are pink.\\
        Reference & Ihre Nagelbetten sind rosa.  \\
        \hline
        Sent-level cur &  Ihre Nagelbetten sind \textbf{pink}. \\
        Doc-level cur & Ihre Nagelbetten sind \textbf{rosa}.\\
        \hline
        \end{tabular}
        \caption{One improved example in OpenSubtitles that is not easily interpretable.}
        \label{tab:example-opensubtitles-na}
 \end{table}

 \begin{table}[ht] 
        \centering
        \small 
        \begin{tabular}{@{} l | l @{}}
        \hline 
        Context & BOUNDARY\_TOKEN \\ 
        Source & We 're making the city liveable.\\
        Reference & Wir machen die Stadt lebenswert.  \\
        \hline
        Sent-level cur &  Wir stellen die lebensfähige Stadt. \\
        Doc-level cur & Wir machen die Stadt lebensfähig. \\
        \hline
        \hline 
        Context & BOUNDARY\_TOKEN \\ 
        Source & Tumlingen is celebrating 750 years \\ 
        & of the St. Hilarius Church.\\
        Reference & Tumlingen feiert 750 Jahre St. \\
        & Hilarius-Kirche. \\
        \hline
        Sent-level cur & Tumlingen ist die Feier von \\
        & 750 Jahren der St. Hilarius Kirche. \\
        Doc-level cur & Tumlingen feiert 750 Jahre der \\
        & St. Hilarius Kirche. \\
        \hline
        \hline 
        Context & BOUNDARY\_TOKEN \\ 
        Source & Prince Henrik does not want a  \\
        & grave next to his wife. \\
        Reference & Prinz Henrik will kein Grab \\
        & neben seiner Frau.  \\
        \hline
        Sent-level cur &  Prinz Henrik will keine große \\
        &Verbindung zu seiner Frau. \\
        Doc-level cur & Prinz Henrik will kein Grab\\
        & neben seiner Frau. \\
        \hline
        \hline 
        Context & BOUNDARY\_TOKEN \\ 
        Source & Pasture fence project is fundamental. \\
        Reference & Weidezaunprojekt ist elementar. \\
        \hline
        Sent-level cur &  Grundlegendes Projekt zur \\
            & Wachstumsfoerderung ist das Projekt. \\
        Doc-level cur & Ein grundlegendes Projekt.\\
        \hline
        \hline
        Context & BOUNDARY\_TOKEN \\ 
        Source & Collision with a car : Cyclist \\
        &hurled onto the road. \\
        Reference & Kollision mit Auto: Radler  \\
        & wirdauf Straße geschleudert. \\
        \hline
        Sent-level cur &  Die Kollision mit einem \\
        &Auto: der Zyniker härtete sich  \\
        & auf die Straße. \\
        Doc-level cur & Die Kollision mit einem Auto. \\
        \hline
        \end{tabular}
        \caption{Example translations in WMT generated by sentence-level baseline and the best document-level mode, i.e., Multi-Encoders (In. Par.).}
        \label{tab:exampe-wmt}
    \end{table}
Table \ref{tab:exampe-wmt} gives some examples of news-headlines translation in WMT where the context is just a special token denoting that the next sentence should be a headline. As clearly seen, the translations generated by the context-aware model are generally shorter and in headline-style as expected.


\end{document}